\documentclass{IOS-Book-Article}     
\usepackage{mathptmx}
\usepackage[T1]{fontenc}
\usepackage{times}
\usepackage{verbatim}
\usepackage{graphicx}
\usepackage{cite}
\usepackage{nameref,hyperref}
\usepackage{mathptmx}
\usepackage{soul}\setuldepth{article}

\def\hb{\hbox to 11.5 cm{}}

\begin{document}

\pagestyle{headings}
\def\thepage{}

\begin{frontmatter}    
\title{The use of Synthetic Data to solve the scalability and data availability problems in Smart City Digital Twins }

\author[A]{\fnms{Esteve} \snm{Almirall}%
\thanks{Corresponding Author: Esteve Almirall, Esade Business School, URL University; E-mail:
esteve.almirall@esade.edu}},
\author[A]{\fnms{Davide} \snm{Callegaro}},
\author[A]{\fnms{Peter} \snm{Bruins}},
\author[B]{\fnms{Mar} \snm{Santamaría}},
\author[B]{\fnms{Pablo} \snm{Martrínez}}
and
\author[C]{\fnms{Ulises} \snm{Cortés}}

\runningauthor{E. Almirall et al.}
\address[A]{Esade Business School, URL University}
\address[B]{300.000km, 300000kms.net}
\address[C]{Knowledge Engineering and Machine Learning Group, Universitat Politècnica de Catalunya}

\begin{abstract}
The A.I. disruption and the need to compete on innovation are impacting cities that have an increasing necessity to become innovation hotspots. However, without proven solutions, experimentation, often unsuccessful, is needed. But experimentation in cities has many undesirable effects not only for its citizens but also reputational if unsuccessful. Digital Twins, so popular in other areas, seem like a promising way to expand experimentation proposals but in simulated environments, translating only the “half-baked” ones, the ones with higher probability of success, to real environments and therefore minimizing risks. However, Digital Twins are data intensive and need highly localized data, making them difficult to scale, particularly to small cities, and with the high cost associated to data collection. We present an alternative based on synthetic data that given some conditions, quite common in Smart Cities, can solve these two problems together with a proof-of-concept based on NO$_2$ pollution.  
\end{abstract}


\end{frontmatter}
\markboth{May 2022\hb}{May 2022\hb}

\section{Introduction}

In 2011 Marc Andreessen wrote a seminal article in the Wall Street Journal, “Software is eating the World.” Today we can argue without exaggeration that Artificial Intelligence is eating software. This is increasingly true in cities. We have and are assisting in the process of city sensorization that brought in an abundance of data.

Autonomous systems are being replaced by centralized command and control systems in areas as diverse as traffic management, garbage collection, pollution control, or even watering gardens. Satellites, drones, vision systems, and many others will undoubtedly increase this abundance and availability of real-time data. 

A consequence of this abundance of data realizes the old dream of a \textit{City brain\/} taking shape. A City Brain is a centralized command and control structure that could optimize everyday life and adapt seamlessly to unexpected situations. This is, however, not the only consequence. We can also witness the increasing use of data for planning to evaluate potential scenarios and proposals. To the extreme, some talk about the redefinition of existing disciplines such as urbanism with urban data and data offices increasing in almost every major city.


Digital Twins have been evolving in aerospace exploration, manufacturing, and many other areas and, together with them, has been an evolution of their understanding, from models of specific characteristics of an artifact to digital shadows closely reflecting it or digital twins that allow the direct manipulation of the artifact through the digital twin. We can observe that the definition is evolving slowly in the Smart City community. Not only because a city is way more complex than a machine or an engineering artifact but also because planning, exploration, and deliberation are at the core of city management.


Two challenges that Digital Twins find in cities are the lack of complete data, particularly real-time data, and the need for scalability. Cities are large, grow and change constantly and have fuzzy borders. The idea of sensorizing a whole city is undoubtedly bold, difficult to attain, challenging to make it economically sound, and even more to keep it updated. In addition to that, cities are similar but also diverse. Thinking of an \textit{ad-hoc} development of digital twins for each town on our planet is unthinkable.


We can observe that in many cities, we face discrete decisions, such as allowing truck traffic or not, let it in defined schedules (\textit{e.g. }incl. or excl. school hours, weekends, …) or even closing the streets. Given this lack of granularity of the decisions, the precision of the digital twin must be tempered by its feasibility and cost. 

In this paper, we introduce “Deep Air,” a proof-of-concept prototype to provide some light on solving these two problems using machine learning and synthetic data. In synthesis, our prototype is a digital twin for city pollution built with synthetic data created by a calibrated machine learning model. We show that the accuracy of the prototype is enough for investigating pollution city policies and reactions to specific conditions, taking into consideration the low granularity of applicable procedures.

This proof-of-concept opens the door to a new generation of digital twins for cities using mixed data from real-time sensors and synthetic models. This approach will not only solve some of the problems of Smart City digital twins but also enable a model of decentralized self-regulated governance based on AI and synthetic data digital twins that we believe are endowed with higher agility, flexibility, scalability, and far lower cost while enabling Open, data-driven innovation in cities.  
\subsection{Plan of the paper}
The paper is organized as follows. Section \S\ref{sec:mat} will discuss digital twins, their presence in Smart Cities, their challenges and limitations, and our proposed framework with its boundary conditions. Section \S\ref{sec:results} will be devoted to the description of the model and discuss the lessons learned with it. Section \S\ref{sec:discussion} will explore the possibilities of this new type of distributed architecture in terms of governance and innovation. And finally, in \S\ref{sec:conclusions} we will conclude.

\section{Materials and Methods}\label{sec:mat}
\subsection{Digital Twins in Smart Cities }
A digital twin is a virtual representation of the characteristics and behaviors of a physical entity used to study and predict its conduct without having to experiment with the actual object \cite{jones2020characterising}. The concept of creating a digital duplicate of a system was practiced at NASA starting from the ’60s where one of the first accounts that we have of this practice was a model of Apollo 13 aimed at simulating the conditions on board in real-time \cite{boschert2016digital},\cite{enders2019dimensions}, aimed at avoiding the failure of the mission. The first crystallization of the concept was in the works of Michael Grieves, and Joan Vickers at NASA in 2003 \cite{grieves2017digital}. It was redefined at NASA in 2012 as an integrated multiphysics, multiscale, probabilistic simulation of a system, which mirrors the life of a corresponding physical artifact based on historical data, physical model, and real-time sensing \cite{glaessgen2012digital}.

Since then, the concept of the digital twin has become popular \cite{tao2018digital}, although with a diversity of approaches and understandings. These different understandings go from mimicking the complete functioning of a physical object, including the capacity to manipulate it through the digital twin, to a broader sense that emphasizes the ability to predict the future state of a system using AI and simulations \cite{cioara2021overview}. 


This diversity of understandings has, among other causes, its roots in the variety of applications of digital twins, being prevalent in manufacturing \cite{malik2021framework},\cite{rovzanec2021actionable} because of its capacity and need for simulations. Similarly, they are even more common in the aerospace industry \cite{rios2015product} because of the extreme cost difficulty of testing, even prototypes. Also, in the automotive industry \cite{chen2018digital}, in smart grid studies \cite{cioara2021overview} or even in socio-technical industries such as manufacturing plants \cite{park2018challenges}.

In general, as closely as the use of a digital twin is to a concrete physical object, the more precise its representation is. At the same time, when it applies to broader systems, particularly socio-economic and human interactions, they lean towards prediction and simulation more.

The case of using digital twins to model Smart Cities lies at the extreme. Cities are one of the most complex constructions of humankind, involving socio-economical, physical, urban, and human interactions. Digital twins in Smart Cities aim at modeling a fraction of these complexities through time and scale, allowing the experimentation of complex policies and their resulting behaviors in cities, something otherwise impossible \cite{francisco2020smart}. 

Smart Cities is probably today one of the most substantial areas of development of digital twins, not only in terms of projects being developed across the world in cities including Barcelona\cite{Graells2021}, Singapore, Helsinki, Boston, Glasgow, and many others, but in terms of broadening our understanding of what a digital twin can be. 


Widespread experiments are on mobility and traffic behaviour under different scenarios and conditions \cite{dembski2019digital},\cite{Graells2021}. However, more common scenarios involve planning (see \cite{Portland17}). Integrating traffic regulation systems such as City Brain in China facilitates the construction and use of these digital twins even more. On the same line, congestion pricing has also been popular because many cities must implement these systems to meet 2030 environmental commitments. 


A systematic analysis of the literature on digital twins in Smart Cities reveals that applications relate to five different themes. Those are visualization, planning and predictions, situational awareness, and data management \cite{shahat2021city}. They also include four significant characteristics \cite{deren2021smart}, mapping (modeling physical aspects of the city), virtual-real interaction (observation in the virtual environment of aspects of the physical one), software models (used for simulations and the own digital twin) and intelligent feedback (the analysis of policy effects either improving cities or potential signaling dangers) \cite{papyshev2021exploring}. 

However, this is a moving target, where the advances in sensors, AI, modeling, and systems integration in cities and the confluence of virtual and physical realities continuously enlarge the space of possibilities and the potential use of digital twins.

Nevertheless, a primary constraint and trigger have been the massive sensorization of urban environments with IoT, the widespread connectivity in cities through 4/5G and Wifi, and the increased availability of detailed satellite imaging. All this has created a data abundance that, for many, precludes a paradigm shift in urban planning and city management where data is going to play a crucial role, and digital twins are poised to be the new wave of urban experimentation tools.


\subsection{The four challenges and limitations of Smart City Digital Twins }

Historically, the dynamics of cities were understood with relatively high levels of abstraction from the perspective of its most extensive infrastructure and their significant impact on sustainability \cite{mohammadi2021thinking}. Smart City digital twins heavily depend on an abundance of data, particularly real-time data, with fine granularity. This abundance of data does not correspond not only to the actual availability of sensors in cities but also to their foreseen implementation. Also, having the data does not imply being able to discover knowledge.

The availability of sensors is undoubtedly one of the most apparent limitations in digital twins in cities, but not the only one. For example, Los Angeles reports that daily, in 2019, they process 7.4 Terabytes of real-time data for four million citizens. Extracted from 4700 traffic signals, 23800 traffic sensors and 1500 bus signals \cite{LADOT19}. \\

This section will discuss three more challenges besides the Data Challenge, the Updating Challenge, the Scalability Challenge, and the Real-time response Challenge. 

\begin{figure}[h]
\includegraphics[scale=0.4]{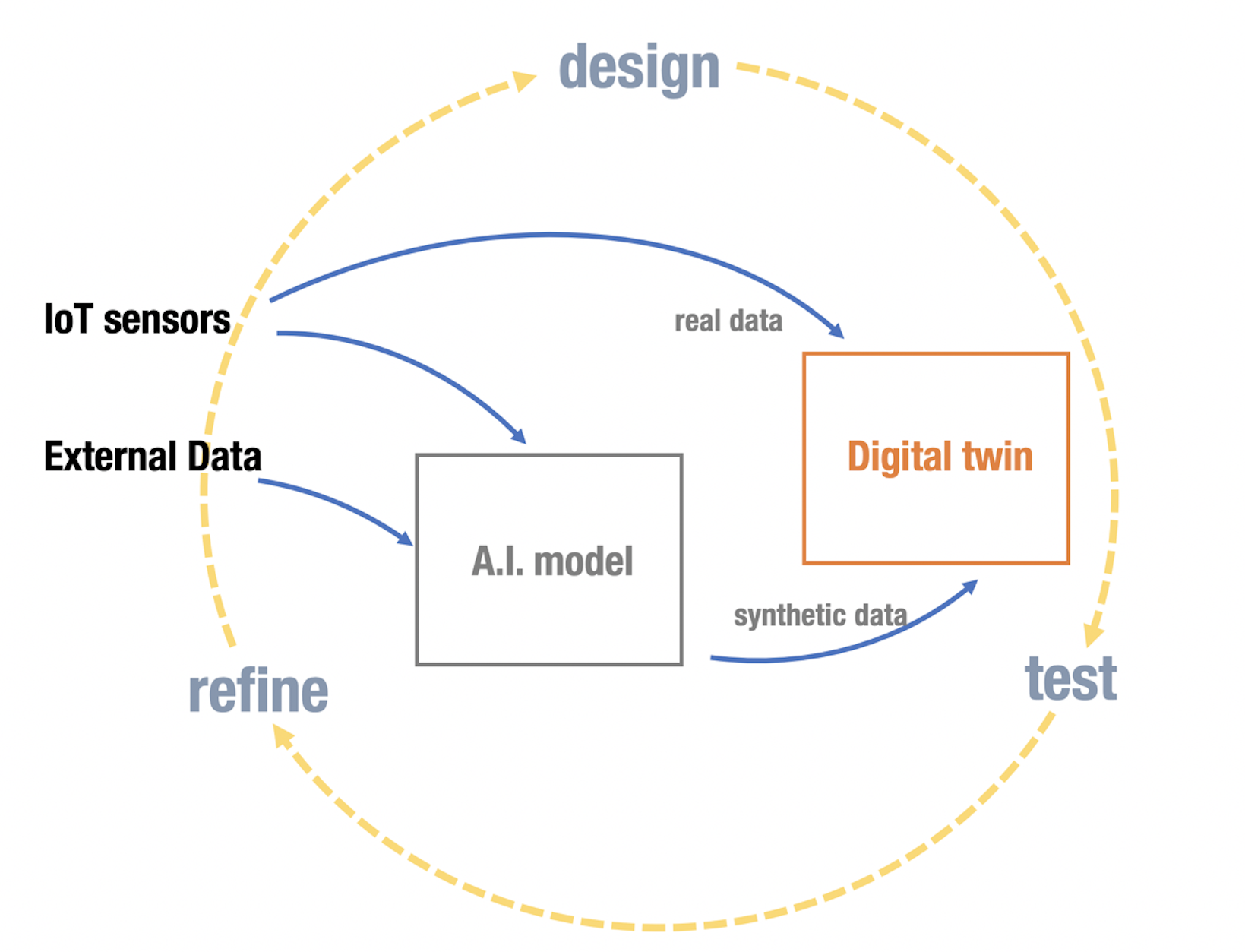}
\caption{A digital twin fed with real and synthetic data in a design-test-refine loop.\label{fig:DATfed}}
\end{figure}

The central idea of our proposal is to use synthetic data. Therefore, data is created through an intermediary A.I.-based model to feed the digital twin together with real data from sensors and synthetic data (see Fig \ref{fig:DATfed}).

Therefore, both data coming from sensors and external data will be the input of a machine learning model that will produce synthetic data for the digital twin. Again, this data and accurate data will feed the digital twin. 

This model requires a loop of designing-test-refining for both the AI model and the digital twin to detect and correct data, model, or concept drifting. This is particularly important because the AI model introduces one more layer and, therefore, one more source of drifting and potential artifacts in the process. 

The use of synthetic data will solve the problem of the scarcity of data. However, it could create another issue, the one of accuracy. We argue, however, that if the accuracy of the AI model supporting the digital twin is high enough and the granularity of the planning decision to be taken based on the data of the digital twin low enough, then there is a space where it will be no difference in terms of decisions between synthetic and real data.
To express it formally, the equality of decision proposition must be satisfied. \\

\noindent \textbf{Proposition 1.} Equality of decisions.\\

Given a set of real data (measurements) \begin{math} \Phi  = \{\phi_1, \phi_2, \phi_3 \ldots \}\end{math} and a set of synthetic data (generated measurements) \begin{math} \hat{\Phi}  = \{ \hat{\phi{_1}}, \hat{\phi{_2}}, \hat{\phi{_3}} \ldots \} \end{math} and given a digital twin \begin{math}T (\Phi)\end{math}, where \begin{math}T (\cdot)\end{math} is a function assuming \begin{math} {\rm I\!R}^{\infty} \end{math} values, we can define the enabling decision function as \begin{math}D(T)\end{math} over a discrete space of decisions \begin{math}D(T)=\{\delta_1, \delta_2, \delta_3, \ldots\}\end{math}. 

Thus, the equality of decisions proposition is satisfied when $D$($T$($\Phi$)) = $D$($T$($\hat{\Phi}$)), and this equality holds if and only if two conditions are met:
\begin{enumerate}
    \item $\Phi $ $\cong $ $\hat{\Phi}$, therefore, a highly accurate model producing synthetic data is needed
    \item $D$ = \{$\delta_1$, $\delta_2$, $\delta_3,$ $\ldots$\} met when $\delta_i$ is highly separated from $\delta_j$, $\forall_{i,j}$
\end{enumerate}

\medskip
Of these two conditions, one is potentially under control, the accuracy of the model, while the other is case-dependent. On some occasions, decisions are significantly separated, for example, the case where the decision is allowing or not allowing truck circulation through a particular street. This will probably be the case in many public policy decisions related to physical infrastructure. In other cases, however, granularity will be more refined, such as the case of congestion taxes, requiring a higher accuracy of the model. 

In this paper, we will concentrate on the first condition 1), providing a proof-of-concept and showing that high accuracy (more than 88\%) is possible in pretty complicated models (NO$_2$ congestion) with a minimal set of variables (eight in this case).

So far, we discussed data availability, which is the major challenge that faces today’s digital twins because increasing data availability implies increasing sensorization, which is not only costly, but lengthy in terms of deployment and, many times, difficult to substantiate in economic terms because of the low interest of remote sensors. However, this is not the only challenge around data. Data updating is also a significant challenge, particularly if real-time data is needed. Data updating, however, can be solved with an accurate machine learning model that provides synthetic data according to real, forecasted, or experimental conditions. In addition to that, a temporal model could fill the void between temporal sensor data. 

Scalability is possibly one of the significant challenges that digital twins face. Developing and calibrating a digital twin is costly and requires not only the installation of sensors but the development of local procedures that automate data collection. 

Again, using a machine learning model as a first step can solve most of these problems. In the proof-of-concept that we present in the coming sections, we will show how a simple model based on external data and existing urban planning data can provide an accuracy high enough to enable a digital twin. Our system will provide accurate enough information leading to decisions indistinguishable from those that could have been taken from real data. The use of synthetic data together with existing and available features enables the high and fast scalability of the digital twin.

Finally, this model covers one more challenge: this is the real-time response challenge. Indeed, all or part of the sensors could be unavailable because of a natural disaster or simply because of an interruption of service. Data from these sensors can be easily covered by synthetic data built by the machine learning model. 

In the \S\ref{sec: results} we will show with a proof-of-concept that this is the case in many situations in Smart Cities because they are driven by physical and urban constraints such as the urban structure.

\section{Results}\label{sec:results}

\subsection{Deep Air – Model Description}

Pollution is one of the most critical problems that cities face. The influence of our lifestyle and urban structure on city pollution has been underlined during the COVID19 crisis. In Barcelona, for example, the NO$_2$ levels dropped to unreachable levels for a long time; compared to everyday life, NO$_2$ levels were roughly 64\% lower in March 2021. Multiple studies have shown that NO$_2$ pollution is associated with various diseases such as diabetes mellitus, hypertension, stroke, chronic obstructive pulmonary disease (COPD), and asthma.

However, NO$_2$ pollution is contingent on many factors, many of them very local and time-dependent, such as traffic density, type of traffic, width of the streets, and, in general, the shape of an urban fabric built through centuries. The challenge to be addressed in this proof-of-concept model is, therefore to what extent is it feasible to create a model with limited data and enough accuracy to feed a digital twin for policy experimentation. 

This is undoubtedly difficult because of data locality, but this case can be generalized to other fields such as different types of pollution, mobility, congestion pricing, etc.

Our results show, however, that it is possible to predict NO$_2$ pollution data with an accuracy of 88.876\%, std of 1.3768, with an XGBoost model primarily based on geographical data and only eight features. These are certainly encouraging results. 

Air pollution is a structural problem that sometimes can be episodic. Today there are several ways to measure air pollution. Air pollution can be described through air stations - points with high accuracy - or satellites - e.g., S5p/troponin with 5 km resolution. It can also be characterized through simulations or Land Use Regressions models (LURs) - e.g., Lobelia/isglobal. However, all these methods have drawbacks in terms of low resolution (satellites), zonal indicators (air stations), or lack the urban complexity, which is a fundamental factor (LURs).
For this project, we have extracted data from multiple sources. Some data sets were publicly available, and others were given upon request.

\begin{itemize}
    \item INE (Instituto Nacional de Estadística) data is all publicly available and found on their website 
    \cite{INE2011}.
    \item EEA (European Environmental Agency) data is publicly available and found on their website \cite{EEA2019}.
    \item 	WAQI (World Quality Air Index) data is publicly available and found on their website \cite{WAQI2020}. 
    \item The datasets held by 300000 km/s (300000kms.net) - an urban think tank located in Barcelona, Spain, which has been part of the research by providing industry insights and ad-hoc data - are private but not confidential and available upon request.
\end{itemize}


All datasets required some cleaning and pre-processing, with changes in the coordinate system to plain coordinates instead of spherical. The first dataset – INE – contains very granular information on income divided by area, population by location and age, dwelling composition (building binned by square meters), and surface division (a division of Spain in polygons representing areas and surface of each polygon).

The European Environmental Agency dataset (EEA) is a massive pool of air station data collected in 2019 in Spain. Some interesting features we noticed are the altitude at which the station is located, the measurements of O$_3$ and NO$_2$ (the first not used in this project), the time of measurement (a column showing the initial time and one showing the end time), and the location of each station (expressed in two columns, one for latitude and one for longitude). 

In addition to this data, datasets held by 300000kms.net contributed to urban data, building density, and people's movements from one location to another during November 2019. After all the data treatments, we had a dataset showing NO$_2$ data per district (a spatial demarcation roughly similar to postal code).

\begin{figure}[h]
    \centering
    \includegraphics[scale=0.6]{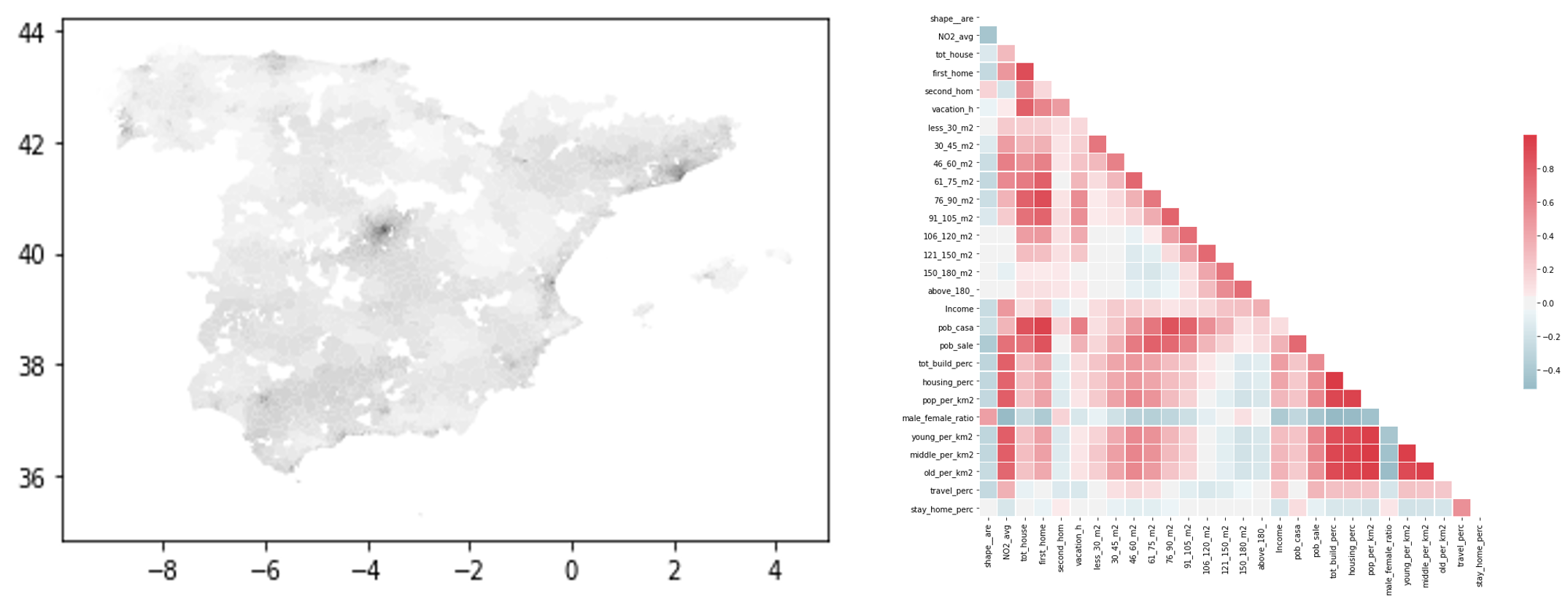}
    \caption{NO$_2$ data per district, together with the correlation matrix of the initial 28 features}
    \label{fig:28feat}
\end{figure}

Initially, dataset collection resulted in approximately one hundred features. Using domain knowledge, an initial selection was made, resulting in a set of 28 elements that constituted our baseline model (Fig. \ref{fig:28feat}).

An initial Random Forest model achieved an accuracy of 85.746\%, std 0.444071\% with an average NO$_2$ level of 12.847696 and an expected error range of [11.856271, 13.839122]).

\begin{figure}[h]
    \centering
    \includegraphics[scale=0.25]{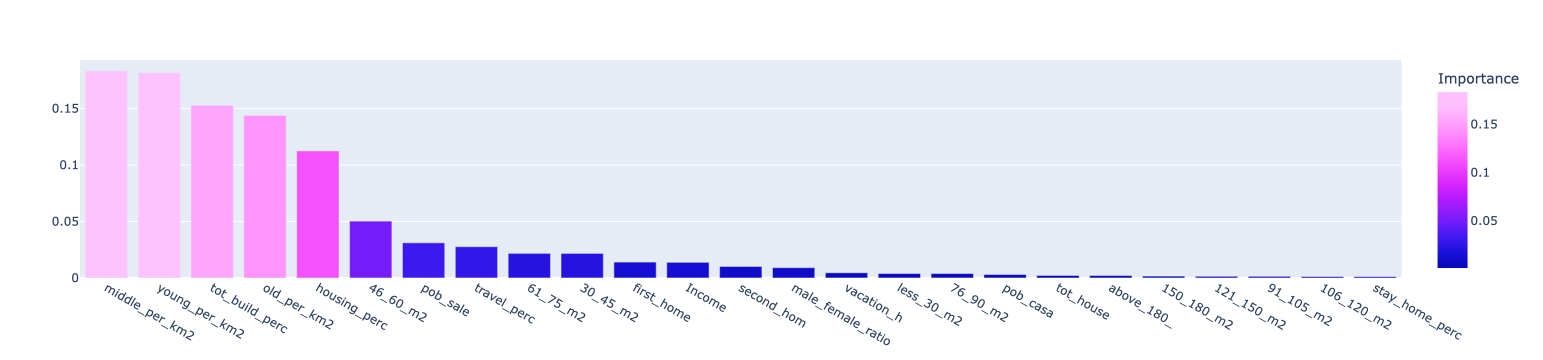}
    \caption{Feature importance - higher is better}
    \label{fig:featImportance}
\end{figure}

However, looking at feature importance, this model shows clear possibilities of simplification in addition to improvement using a more sophisticated algorithm (See Figure \ref{fig:featImportance}).

The Moran's I statistic explains the degree to which a feature has a spatial relation. Zero implies no extra information to be added by using a spatial lag. On the other hand, one means that knowing the neighboring values of a feature gives an almost 100\% certainty of the value of the feature at hand. Later, we would decide that the cut-off point to include a feature as spatial lagged was a Moran I of around 0.6. Values close to 0.6 were tested individually to check for the best performance.

We performed feature selection among the mix of lagged and non-lagged features, reducing them to only eight and using a Random Forest model. We achieved an accuracy of 88.651\%, std 0.979779, with an average NO$_2$ level of 12.847696 and an expected error range of [11.856562, 13.838831]. Below you can observe a graphical presentation of the accuracy of the model.

\begin{figure}[h!]
    \centering
    \includegraphics[scale=0.5]{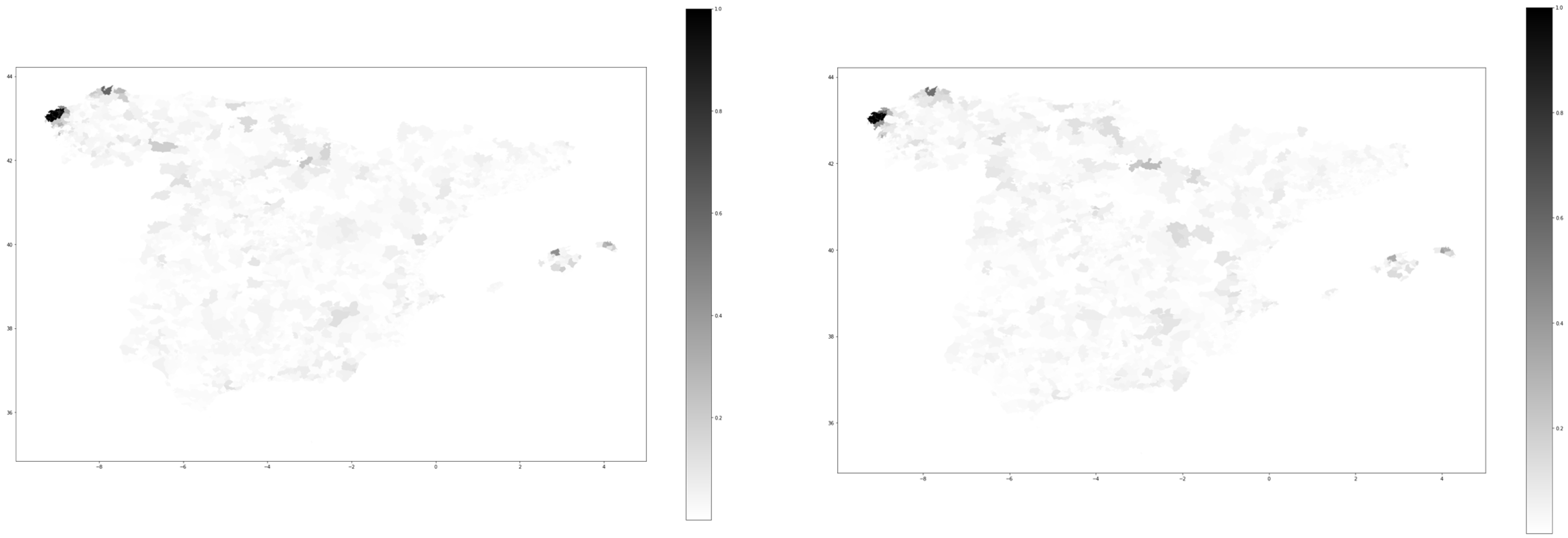}
    \caption{Random Forest and XGBoost model accuracy – darker is less accurate}
    \label{fig:RF_XGB}
\end{figure}

Finally, we reproduce the same analysis with a more powerful model, an XGBoost, obtaining some improvement and reaching our final accuracy of 88.87\%, std 1.376834 (See figure \ref{fig:RF_XGB}).

The model started with more than 100 features that were reduced to a list of thirty, and from them, through feature selection, reduced to eight while increasing the accuracy of the prediction. 

This is only a proof-of-concept that lacks essential elements such as traffic flows. However, we can argue that we could obtain similar results by adding traffic flows and more granular data, for example, at the city level. 

As shown, only a few elements are determinants of NO$_2$ pollution, many of them static because they are part of the urban fabric. Therefore, it is possible to create models with limited features and high accuracy.

The next evolution of this model will undoubtedly need to involve traffic data and put more emphasis on dynamic data. A development proposal could be using a GNN, where edges represent streets or roads and vertices cities. Such a model will provide more flexibility to represent timely based traffic data.

\section{Discussion}\label{sec:discussion}

Through this paper, we have shown how synthetic data is feasible and solves many problems that today’s digital twins face in complex socio-economic environments such as cities.

However, they also open the door to new possibilities. Opening the digital twin by allowing the contribution of many providers and organizations to transform it into an instrument of collaboration and experimentation effectively is also a new frontier that the dual design we proposed facilitates.

Opening digital twins in cities and thus transforming them into an available tool for experimentation will allow the plug of new A.I. data models to provide better or new synthetic or simulated data. But it also will allow the test and validation of new city proposals in a collaborative sandbox environment. 

More than that, its opening will provide the necessary affordabilities to build an ecosystem in two directions. First, further developing the digital twin and, secondly, allowing the deployment of informed proposals that can show its potential not only to one city but to many.

\section{Conclusions}\label{sec:conclusions}

Undoubtedly, the need for better models for digital twins in cities is on the rise. Not only because of the complexity of cities and the growing performance expectations of citizens but also because of the speed and non-linearity of change that cities endure.

For example, cities will face the arrival of autonomous cars in the coming years. After a few years of experimentation by Tesla and Google in Portland, where the service is active, we realized that this would not be a uniform, wide adoption, and rapid phenomenon. Instead, we will find an uneven adoption pattern with various technologies. 
Therefore, more than ever, cities will need tools that can help them experiment with policy alternatives. Cities need new ways to perform the savvier digitization of their policies.

On the other hand, we are and will be far from the complete sensitization of cities, not only because of its feasibility in terms of the time needed to carry out the said project but also in terms of economic viability. 

Digital twins, in their broad sense as digital models that could accurately represent certain aspects of a city, allowing for experimentation, planning, knowledge discovery - metadata will be a valuable asset- and even near real-time adjustments or emergency activation procedures, are part of the future of management, optimization and city planning. Metaphorically we can say that code is the new concrete \cite{LADOT19}.

However, if digital twins use only real data, there are insurmountable barriers to availability, scalability, updating, and real-time response. Their modularization using future synthetic data plugins and opening as external experimentation tools could enable the open experimentation and collaboration that cities need. 
Last but not least, we should also mention the environmental concerns of complete sensitization that could be lessened with this focus in the direction of green algorithms.

The proof of concept present in this work shows an alternative path that mixes real and synthetic data. We hope this research inspires a new generation of digital twins to support cheaper, more scalable, and open city management enabling open, data-driven innovation in cities.

\bibliographystyle{vancouver}
\bibliography{AirDeep}

\begin{thebibliography}{10}

\bibitem{jones2020characterising}
Jones D, Snider C, Nassehi A, Yon J, Hicks B.
\newblock Characterising the Digital Twin: A systematic literature review.
\newblock CIRP Journal of Manufacturing Science and Technology. 2020;29:36-52.

\bibitem{boschert2016digital}
Boschert S, Rosen R.
\newblock Digital twin—the simulation aspect.
\newblock In: Mechatronic futures. Springer; 2016. p. 59-74.

\bibitem{enders2019dimensions}
Enders MR, Ho{\ss}bach N.
\newblock Dimensions of digital twin applications-a literature review. 2019.

\bibitem{grieves2017digital}
Grieves M, Vickers J.
\newblock Digital twin: Mitigating unpredictable, undesirable emergent behavior
  in complex systems.
\newblock In: Transdisciplinary perspectives on complex systems. Springer;
  2017. p. 85-113.

\bibitem{glaessgen2012digital}
Glaessgen E, Stargel D.
\newblock The digital twin paradigm for future NASA and US Air Force vehicles.
\newblock In: Structural dynamics and materials conference 20th AIAA/ASME/AHS
  adaptive structures conference 14th AIAA; 2012. p. 1818.

\bibitem{tao2018digital}
Tao F, Zhang H, Liu A, Nee AY.
\newblock Digital twin in industry: State-of-the-art.
\newblock IEEE Transactions on Industrial Informatics. 2018;15(4):2405-15.

\bibitem{cioara2021overview}
Cioara T, Anghel I, Antal M, Salomie I, Antal C, Ioan AG.
\newblock An Overview of Digital Twins Application Domains in Smart Energy
  Grid.
\newblock arXiv preprint arXiv:210407904. 2021.

\bibitem{malik2021framework}
Malik AA.
\newblock Framework to model virtual factories: a digital twin view.
\newblock arXiv preprint arXiv:210403034. 2021.

\bibitem{rovzanec2021actionable}
Ro{\v{z}}anec JM, Lu J, Rupnik J, {\v{S}}krjanc M, Mladeni{\'c} D, Fortuna B,
  et~al.
\newblock Actionable Cognitive Twins for Decision Making in Manufacturing.
\newblock arXiv preprint arXiv:210312854. 2021.

\bibitem{rios2015product}
R{\'\i}os J, Hern{\'a}ndez JC, Oliva M, Mas F.
\newblock Product avatar as digital counterpart of a physical individual
  product: Literature review and implications in an aircraft.
\newblock Transdisciplinary Lifecycle Analysis of Systems. 2015:657-66.

\bibitem{chen2018digital}
Chen X, Kang E, Shiraishi S, Preciado VM, Jiang Z.
\newblock Digital behavioral twins for safe connected cars.
\newblock In: Proceedings of the 21$^{th}$ ACM/IEEE International Conference on
  Model Driven Engineering Languages and Systems; 2018. p. 144-53.

\bibitem{park2018challenges}
Park H, Easwaran A, Andalam S.
\newblock Challenges in digital twin development for cyber-physical production
  systems.
\newblock Cyber Physical Systems Model-Based Design. 2018:28-48.

\bibitem{francisco2020smart}
Francisco A, Mohammadi N, Taylor JE.
\newblock Smart city digital twin--enabled energy management: Toward real-time
  urban building energy benchmarking.
\newblock Journal of Management in Engineering. 2020;36(2):04019045.

\bibitem{Graells2021}
Graells-Garrido E, Serra-Burriel F, Rowe F, Cucchietti F, Reyes P.
\newblock {A city of cities: Measuring how 15-minutes urban accessibility
  shapes human mobility in Barcelona}.
\newblock PLoS ONE. 2021;16(5):e0250080.

\bibitem{dembski2019digital}
Dembski U Fabian;~Wössner, Yamu C.
\newblock Digital twin, virtual reality and space syntax: Civic engagement and
  decision support for smart sustainable cities.
\newblock In: 12th International Space Syntax Symposium Beijing; 2019. .

\bibitem{Portland17}
{Portland City Hall}. {Portland Plan. Progress report}; 2017.
\newblock Available from: \url{https://www.portlandonline.com/portlandplan/}.

\bibitem{shahat2021city}
Shahat E, Hyun CT, Yeom C.
\newblock City digital twin potentials: A review and research agenda.
\newblock Sustainability. 2021;13(6):3386.

\bibitem{deren2021smart}
Deren L, Wenbo Y, Zhenfeng S.
\newblock Smart city based on digital twins.
\newblock Computational Urban Science. 2021;1(1):1-11.

\bibitem{papyshev2021exploring}
Papyshev G, Yarime M.
\newblock Exploring city digital twins as policy tools: A task-based approach
  to generating synthetic data on urban mobility.
\newblock Data \& Policy. 2021;3.

\bibitem{mohammadi2021thinking}
Mohammadi N, Taylor JE.
\newblock Thinking fast and slow in disaster decision-making with Smart City
  Digital Twins.
\newblock Nature Computational Science. 2021;1(12):771-3.

\bibitem{LADOT19}
{LADOT}. {Los Angeles. Technology Action Plan}; 2019.
\newblock Available from:
  \url{https://ladot.io/wp-content/uploads/2019/03/LADOT-TAP-v7-1.pdf}.

\bibitem{INE2011}
{Instituto Nacional de Estadistica}. {Census Data 2011}; 2011.
\newblock Available from:
  \url{https://www.ine.es/censos2011_datos/cen11_datos_resultados_seccen.htm}.

\bibitem{EEA2019}
{European Environmental Agency}. {Air Pollution Data – EEA}; 2019.
\newblock Available from: \url{https://www.eea.europa.eu/themes/air/dc}.

\bibitem{WAQI2020}
{World Air Quality Index project}. {World Air Quality Index}; 2020.
\newblock Available from: \url{https://www.eea.europa.eu/themes/air/dc}.

\end{thebibliography}


%


\end{document}